\documentclass{article}

\usepackage[final,nonatbib]{cml4impact_2022}

\usepackage[utf8]{inputenc} 
\usepackage[T1]{fontenc}    
\usepackage{hyperref}       
\usepackage{url}            
\usepackage{booktabs}       
\usepackage{amsfonts}       
\usepackage{nicefrac}       
\usepackage{microtype}      
\usepackage{xcolor}         
\usepackage{algorithm}
\usepackage[noend]{algpseudocode}
\usepackage{graphicx,psfrag,epsf}
\usepackage{multirow}

\title{A Novel Two-level Causal Inference Framework for On-road Vehicle Quality Issues Diagnosis}

\author{%
Qian Wang \\
Ford Motor Company\\
Dearborn, MI48124 \\
\texttt{qwang435@gatech.edu} \\
\And
Huanyi Shui\\
Ford Motor Company\\
Dearborn, MI48124 \\
\texttt{hshui@ford.com} \\
\And
Thi Tu Trinh Tran\\
Ford Motor Company\\
Dearborn, MI48124 \\
\texttt{ttran98@ford.com} \\
\And
Milad Zafar nezhad\\
Ford Motor Company\\
Dearborn, MI48124 \\
\texttt{MZAFARNE@ford.com} \\
\And
Devesh Upadhyay\\
Ford Motor Company\\
Dearborn, MI48124 \\
\texttt{dupadhya@ford.com} \\
\AND
Kamran Paynabar\\
Georgia Institute of Technology\\
Atlanta, GA30332\\
\texttt{kamran.paynabar@isye.gatech.edu} \\
\And
Anqi He\\
Ford Motor Company\\
Dearborn, MI48124 \\
\texttt{AHE6@ford.com} \\
}

\usepackage[
backend=biber,
style=alphabetic,
sorting=ynt
]{biblatex}

\begin{document}

\maketitle

\begin{abstract}
    In the automotive industry, the full cycle of managing in-use vehicle quality issues can take weeks to investigate. The process involves isolating root causes, defining and implementing appropriate treatments, and refining treatments if needed. The main pain-point is the lack of a systematic method to identify causal relationships, evaluate treatment effectiveness, and direct the next actionable treatment if the current treatment was deemed ineffective. This paper will show how we leverage causal Machine Learning (ML) to speed up such processes. A real-word data set collected from on-road vehicles will be used to demonstrate the proposed framework.  Open challenges for vehicle quality applications will also be discussed. 
\end{abstract}

\section{Introduction}
On-road quality issues of automobile vehicles not only cause financial burdens for the Original Equipment Manufacturer (OEM) but also impose repair based costs and delays to the vehicle owner. If unchecked these issues can adversely impact customer loyalty. Efficient root cause identification and mitigation can help alleviate some of these impacts.

To the best of our knowledge, an iterative approach for managing on-road vehicle quality issues systematically is not found in literature. In an industrial setting, the process of identifying and eliminating root causes for a given concern can take weeks since domain experts typically investigate for root causes from a large signal corpus, and recommend a fix. If the deployed treatment is found to be ineffective, the cycle repeats until convergence to zero faults. There are mainly three characteristics in data that are bottlenecks for providing solutions for vehicle quality issues efficiently:

1) imbalanced data: as would be expected, the number of reported failure examples are relatively small, typically ranging from 1\%-10\% of the entire vehicle population. 

2) high dimensional data: there are a multitude of factors that can directly or indirectly lead to failures, such as differences in vehicle attributes, vehicle usage, etc.

3) dynamic data set: vehicle usage is not static. If not fully covered in the original data, out-of-distribution changes in these factors can impact the treatment effectiveness analysis. Additionally, there is no guarantee that data before and after the treatment on all vehicles can be collected due to varying levels of user authorization for data connectivity. This can often make assessment difficult.

To overcome these challenges, this paper focuses on developing a causality based analysis framework to improve the efficiency of such find-fix-track process to address on-road vehicle quality issues. The framework mainly consists of two parts - root cause analysis and treatment effectiveness analysis. Traditional root cause analysis - filtering methods focus on correlations but fail to consider interactions between feature and the influence of confounders. Example algorithms include Information Gain \cite{thomas2006elements}, Symmetrical Uncertainty \cite{10.5555/1972514}, Correlation-based Feature Selection \cite{hall1999correlation}, Chi-squared statistic \cite{ugoni1995chi} and Fisher Score \cite{gu2012generalized}. Machine learning based root cause analysis - embedded methods investigate the interactions between features, but still fail to eliminate effects of confounders. Some example algorithms are Random Forest \cite{breiman2001random}, XGBoost \cite{chen2016xgboost}, and Logistic Regression \cite{hosmer2013applied}. In this paper, we propose to employ causal machine learning methods to not only investigate feature interactions, but also eliminate effects from confounders.

The analysis of treatment effectiveness can be conducted using A/B tests given a fully randomized experimental design. However, real-world vehicle data is fully observational, and outcomes can be influenced by many external factors. Therefore, causal inference methods need to be employed to estimate the efficacy of treatment under the influence of confounders. State-of-the-art methods in this area include Propensity Score Matching \cite{rosenbaum1983central}, Inverse Propensity Weighting \cite{robins1994estimation}, Doubly Robust Estimator \cite{bang2005doubly}, Causal Forest \cite{wager2018estimation}, and Meta-learners (for example X-learner \cite{kunzel2019metalearners}). However, these methods only focus on estimation rather than the statistical testing of the significance of effects. In addition, current methods do not provide additional information for improving treatments.

To overcome the above difficulties and fill the gaps between current literature and practical needs, this paper proposes a two-level causal inference framework. The first level identifies the potential causes of a quality issue of interest, calculates the causal effects of any prior treatments applied and builds statistical tests for the significance of treatment effect at both entire population and individual vehicle levels. The second level employs causal inference on top of the causal effect estimation from the first level to interpret the direction and significance of each feature's impact on the treatment. This information can be used to understand the following: given the current treatment, what is the next potential causes to improve the treatment.

The main contribution of this paper is that to our knowledge this may be the first paper in the literature that focuses on iteratively resolving on-road vehicle quality issues by using causal inference methods. A complete framework is proposed for handling the full mitigation cycle including the identification of root causes, the statistical tests on the treatment effectiveness based on fully observational data, and the causal analysis of the critical features for the further improvement of the treatment. Comparing to ad-hoc methods, the proposed framework overcomes the difficulties of considering the interaction between features and confounders using causal inference. Also, this may be the first attempt at implementing an iterative statistical testing procedure for sequential improvement of treatments from observations before and after treatments at both entire vehicle population level and individual vehicle level. Our approach continues to improve treatments until convergence to zero concerns. Finally, we also discuss results from a real-world use case. This use case data is collected from on-road vehicles, is high-dimensional, highly-imbalanced with binary vehicle attribute features, continuous covariates - vehicle usage features and binary outputs.

\section{Methodology}

\begin{figure}
\includegraphics[scale=0.43]{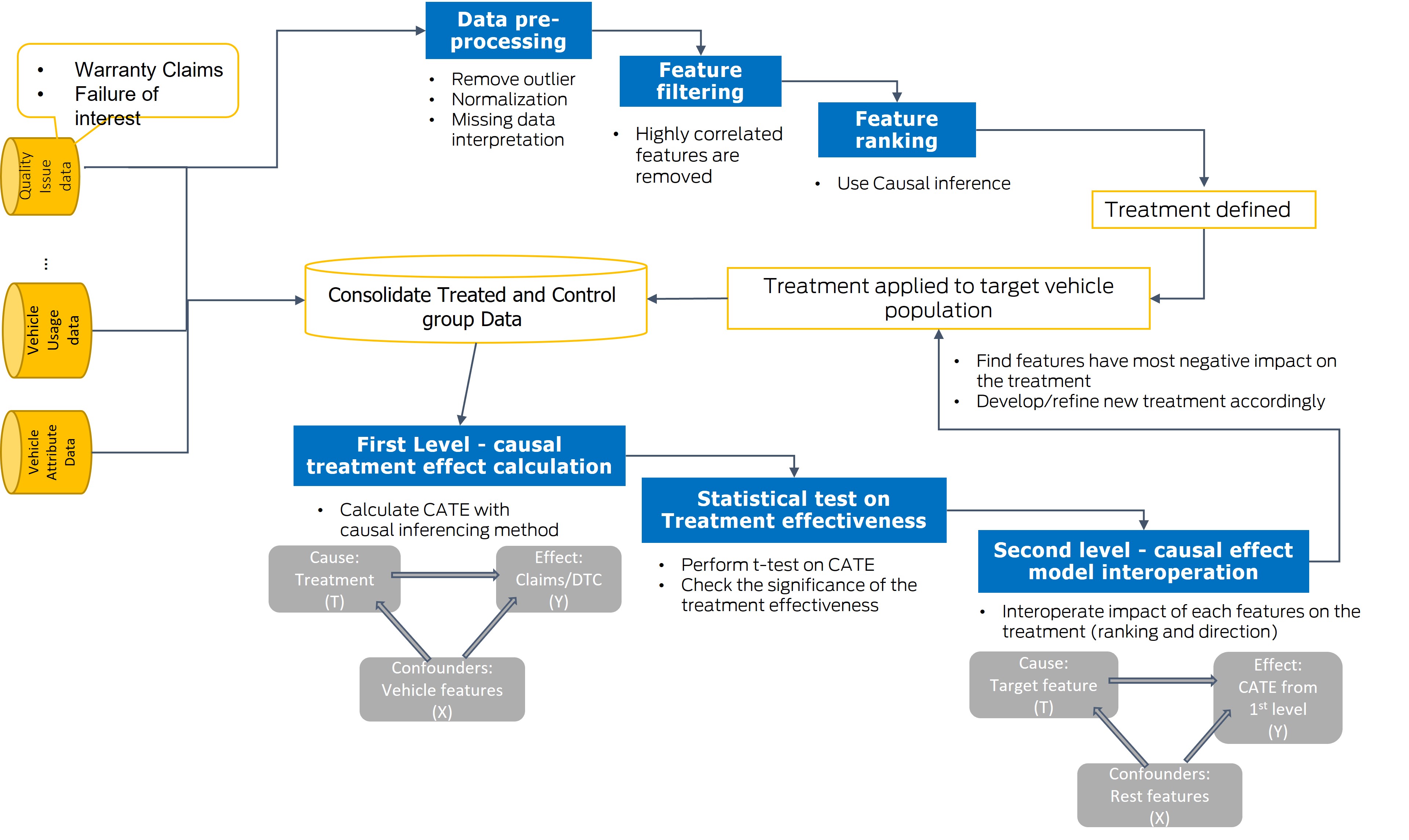}
\centering
\caption{\label{flowchart}Flowchart of the proposed causal inference framework}
\end{figure}

Assume we have data of the form $\{(x^{(k)}, y^{(k)}, T^{(k)}, z^{(k)}), k=1,2,\cdots,N\}$ where $x=(x_1,x_2,x_i,...,x_n)$ are binary features indicating whether a vehicle has the corresponding attribute $x_i$  (can also be referred as vehicle packages or modules). One or more features can be a potential cause for some quality issue; $y=(y_1,y_2,...,y_m)$ are continuous co-variate features representing vehicle usage (i.e. weather, temperature or driving speed) which can influence (confound) the root cause; $T$ is an binary indicator for whether the treatment is applied ($T=1$ if so); and $z$ is a binary response indicating if the vehicle has quality issue under consideration ($z=1$ if so). For an overview of our proposed method, please see the flowchart attached (Figure \ref{flowchart}). The corresponding pseudocode can be found in the Supplementary Material.


\subsection{Feature filtering method for elimination of highly correlated binary features}

Highly correlated features usually exist in industrial datasets, and they influence the performance of the subsequent root cause analysis. This is because correlated features will lead to the split of the same feature importance into each of the highly correlated features thereby removing significance from all features. This calls for the elimination of highly correlated features before further causal analysis. 

This paper develops an empirical method to collect representative features \textit{rep} from the set of  all features and build a mapping to refer back from representatives to original features. Details can be found in the Supplementary Material. For the rest of this paper, input features used for analysis are all representative features, \textit{rep} (i.e. highly correlated features are eliminated).

\subsection{Root cause analysis based on causal inference}
This section introduces the use of causal inference for the analysis of the potential root causes of a quality issue of interest. At this step, we have not yet identified the corresponding treatments, so data is only of the form $(x^{(k)}, y^{(k)}, z^{(k)})$.

First, we train the model $z=f(x,y)$ using XGBoost and remember to balance the class weight of $z$ to compensate for the minority class $z=1$ i.e. classes with quality issues reported/flagged. The quality outcome $z$ varies with different vehicle features $(x,y)$ and the features that drive the quality outcome closest to 1 are considered as strong influencers of the quality issue (since having quality issues means $z=1$). These are the potential root causes that a treatment should focus on. 

Thus, the strategy is to learn a first level model $z = f(x,y)$ using XGBoost to estimate the influence of the vehicle features $(x_1,x_2,..,x_n,y_1,y_2,...,y_m)$ on the quality outcome $z$. After training $f$, the estimation for the effect of vehicle feature $i$ on the quality outcome $z$ is evaluated as follows:
$$
\bar{e}_i=\frac{1}{N}\sum_k (f_{i1}(x^{(k)},y^{(k)})-f_{i0}(x^{(k)},y^{(k)}))
$$
where $$f_{i1}(x,y) = f(x_1,..,x_i=1,...,x_n,y_1,...,y_m)$$ $$f_{i0}(x,y) = f(x_1,..,x_i=0,...,x_n,y_1,...,y_m)$$

Subsequently, the vehicle features with top "m" largest positive $\bar{e}_i$ are suggested as potential root causes leading to positive quality outcome $z=1$. These "m" features are then used as heuristics to develop treatment $T$ accordingly.

In this step, other root cause identification methods, as discussed in the introduction, can be used as well.

\subsection{Statistical testing on treatment effectiveness}
This section discusses how to leverage the data collected before and after treatment for the analysis of treatment effectiveness. The analysis has two steps: first we calculate the treatment effect for each vehicle; second, we build statistical tests for treatment effects at both the population level and at individual vehicle level.

Here we would like to explain the data collection before introducing the analysis methods. To test if the treatment is effective, we form two groups, the treatment group and the control group. The treatment group contains vehicles that have received the treatment, while the control group contains vehicles that have not yet received the treatment at the time of assessment. In both treatment and control groups, data is collected before and after the treatment application. Each group includes data from both problematic and healthy vehicles. After applying the treatment, the data has the form $(x^{(k)},y^{(k)},T^{(k)},z^{(k)})$. 

To analyze the treatment effectiveness, standard causal inference method like meta-learners (this paper uses X-learner) is applied to the data and the Conditional Average Treatment Effect (CATE) of each vehicle is calculated as $e_T^{(k)}$. During the training of base learners, since the quality outcome of vehicles may be biased, classes should be balanced. In this study, more weight is put on the minority class $z=1$ i.e. the problematic ones. 

To develop statistical tests to check the significance of the treatment effectiveness on the vehicle population, we assume CATE over the population to have Gaussian distribution $N\left(\mu, \sigma^{2}\right)$ with Average Treatment Effect (ATE) as the population mean $\mu$. For a treatment to be effective at population level is equivalent to $ATE<0$, and the hypothesis testing can be set up as 
$$
H_{0}: \mu \geq 0 ;\ \ 
H_{1}: \mu<0
$$

T test is the standard way to test the sign of the population mean \(\mu\).
Let \(\bar{d}:\) sample mean, \(s_{d}\): sample standard deviation, $N$: number of samples, then
$$
\mathrm{t}=\sqrt{N} \bar{d} / s_{d}
$$
We can declare ATE \(=\mu<0\) i.e. the treatment is significantly effective over the population at level \(\alpha\) if
$$
t<t_{N-1, \alpha}
$$

However, it is trickier to assess if the treatment is effective for a specific vehicle. This can be informative in diagnosing the root cause. We apply bootstrapping to achieve a finer-level information of treatment effectiveness. A treatment being effective at the individual vehicle level (\(x^{(k)}, y^{(k)}\)) is equivalent to testing for \(CATE \left( x^{(k)}, y^{(k)}\right)<0\). We can apply \(\mathrm{X}\)-learner on 1000 bags of bootstrapped data to calculate a two-sided \(1-2 \alpha\) confidence interval of \(\mu^{(k)}:= CATE \left(x^{(k)}, y^{(k)}\right)\), say $[lb,ub]$, which means

$$
\mathrm{P}(l b \leq \mu^{(k)} \leq u b)=1-2 \alpha \Rightarrow \mathrm{P}(u b<\mu^{(k)})=\alpha
$$
We can declare \(\mu^{(k)}=CATE\left(x^{(k)}, y^{(k)}\right)<0\) i.e. the treatment is significantly effective on the individual vehicle (\(x^{(k)}, y^{(k)}\)) at level \(\alpha\) if
$$
u b<0
$$

\subsection{Improvement analysis based on causal inference}
After the hypothesis testing, a second level causal inference is employed to understand the effect of different vehicle features on the treatment, which can help the investigation on - which vehicle feature should be targeted next for further treatment improvement, if needed.

The treatment effect $e_T^{(k)}$ varies with different vehicles. The vehicle features that influence $e_T$ in the negative direction the most should be the one that the current treatment has addressed, as this tends to move $z$ most negatively from 1 to 0. On the other hand, the features that affected $e_T$ in the most positive direction should be the next candidate for improvement via treatment, as it tends to move $z$ most positively from 0 to 1. 

Thus, a second level causal effect model $e_T = h(x,y)$ is built using XGBoost to estimate the influence of different vehicle features $(x_1,x_2,..,x_n,y_1,y_2,...,y_m)$ on the treatment effect CATE $e_T$, which is calculated at the first level for each vehicle. After $h$ is trained, the estimation formula for the effect of vehicle features $i$ on the treatment effect $e_T$ is 
$$
\bar{e}_i^{2nd}=\frac{1}{N}\sum_k (h_{i1}(x^{(k)},y^{(k)})-h_{i0}(x^{(k)},y^{(k)}))
$$
where $$h_{i1}(x,y) = h(x_1,..,x_i=1,...,x_n,y_1,...,y_m)$$ $$h_{i0}(x,y) = h(x_1,..,x_i=0,...,x_n,y_1,...,y_m)$$


Subsequently, the vehicle features with top "m" largest positive $\bar{e}_i^{2nd}$ are selected. Since these are the major features that lead to the ineffectiveness of the current treatment $T$, we should use them as heuristics to develop a new treatment $T'$. This cycle can be iterated until a satisfactory treatment for the quality issues is developed. This concludes all analytical discussions of the proposed framework.

\section{Use case study}
This section discusses the application of the proposed framework on a  real-world use case for a vehicle fleet. The problem at hand was related to high battery drain after vehicle key-off \cite{battery}. This issue lead to dead batteries, sometimes requiring early battery replacement under warranty. To fix the battery drain issue and avoid additional warranty cost, engineers leveraged connected vehicles data and the proposed framework to identify the root causes, take action, track the treatment effectiveness, and refine the treatments as necessary.

Overall, the data consists of three parts. First, input $X,Y$ includes binary features for vehicle attribute and continuous features for vehicle usage. The total number of features are greater than 1500. Second, the output $Z$ indicates the existence of vehicle battery drain issue. The number of problem cases was less than 10\% of the total fleet population. The last component of data set is a treatment label, $T$, indicating whether a vehicle had a successful treatment applied. Treatments were made available via over the air updates (OTA). Since successful OTA updates can vary by vehicle and availability of connectivity, this provides an opportunity for a natural clustering of a treatment group and a control group. Hence, the control group forms with vehicles that have not yet had the treatment at the time of assessment.

The root cause for the battery drain issue was found to be related to a particular vehicle attribute (Feature A) in a particular model year vehicle. Therefore, the entire population is targeted at this model year vehicles. The treatment was then applied via OTA. Further sampling showed that approximately 75\% of the impacted vehicles had successfully received this treatment. The remaining vehicles naturally formed the control group. Even though we tried to collect data for both before and after treatment, there still remained less than 10\% of vehicles that lacked either before-treatment or after-treatment data due to inconsistencies in connectivity authorization by users. It is noted that the binary features - vehicle attributes are static, while the continuous features - vehicle usage might change overtime, i.e. for a same vehicle, before and after treatment, the distribution of continuous features might be changed.

All codes are built in Python, with major packages from CausalML \cite{chen2020causalml} and EconML \cite{econml}. To clean and filter the data, all missing data is interpreted by filling with median values, and representative features are selected, which reduces the dimension of feature space from 1500+ to around 300. The root cause analysis using the causal ML approach shows that the Feature A is within top 10 risk factors. 

For treatment analysis, first, to evaluate the effect of mix types features, we evaluated the proposed framework with two different scenarios. One is with only binary features $x$, and the other with both binary $x$ and continuous features $y$. 

Second, to calculate the causal treatment effect (CATE), four different causal effect models are built based on different methods - 1) Meta X-learner with Logistic Regression (LR) for propensity score calculation and XGBoost for CATE, 2) Meta X-learner with XGBoost for both propensity score and CATE calculation, 3) Causal Forest - Double Machine Learning (CF-DML) \cite{chernozhukov2018double} and 4) Causal Forest - Generalized Random Forest (CF-GRF) \cite{athey2019generalized} for CATE. 

Then the proposed statistical test is conducted on the calculated CATE. As shown in Table\ref{res}, given the critical value $t_{N-1, \alpha}= -1.645$  at level $\alpha = 0.05$, all test statistics are much smaller than the critical value, which shows the current treatment $T$ is significantly effective for eliminating the battery drain issue under all models. 

Finally, another three approaches are stacked with each learnt first-level causal effect model to interpret the influence direction and significance of each vehicle feature on the treatment effect. The three methods are 1) SHAP \cite{lundberg2017unified}, 2) the proposed $2^{nd}$-level Meta S-learner (Meta S), 3) self-interpretation of tree based methods (Self, only for Causal Forest). The first two methods can provide both direction and significance information, while the third method can only provide the significance. For a given vehicle binary feature \textit{A} as the known root cause and knowledge of the target of the current treatment, we evaluated the performance of the proposed framework by comparing its capability for interpreting the effect of this particular feature on the treatment. In other words, results should show that this feature has a strong impact on the treatment effectiveness. During the evaluation, we combine the direction and significance information into \textit{Ranking in the good direction} as the evaluation metric for the first two methods, while only use significance information for ranking for the third. Feature \textit{A} should be ranked on top for all three methods. The comparison results are shown in Table \ref{res}. 

First, we found that the ranking results from the models built with only binary features outperform those with both binary and continuous features. It it noted that we only give ranks for binary features. This indicates that mix types of input features may challenge the base-model learning.

Second, self-interpretation fails to give direction information, and this makes it incapable of identifying features for treatment improvement as we cannot distinguish "good" features from "bad" features. Also, it is only available for Causal Forest (tree based) type of methods and not generally applicable.

When stacking SHAP with Causal Forest methods, it takes significantly longer to interpret the model. Hence SHAP is not suitable for tree based methods when time matters. But when stacked with Meta Learners, it performs well and gave top rankings for the target feature. Meta Learner with LR+XGBoost performs better than Meta learner with XGBoost only.

Finally, our proposed second level S-learner is generally stackable with any first-level causal model and can interpret the results in a reasonable amount of time (minutes) and gave top rankings with direction information. Moreover, the second level S-learner results in higher ranking when stacked with Causal Forest models than Meta Learners.

\begin{table}[]
\caption{Comparison results between four different causal models under two scenarios}
  \label{res}
  \centering
\begin{tabular}{cccccccccc}
\hline
\multirow{3}{*}{CATE model} & \multicolumn{4}{c}{Binary only}                                      &  & \multicolumn{4}{c}{Binary + Continuous}                              \\ \cline{2-5} \cline{7-10} 
                            & \multirow{2}{*}{t-test stat} & \multicolumn{3}{c}{feature interpretation} &  & \multirow{2}{*}{t-test stat} & \multicolumn{3}{c}{feature interpretation} \\ \cline{3-5} \cline{8-10} 
                            &                         & SHAP           & Meta S       & Self       &  &                         & SHAP         & Meta S        & Self        \\ \cline{1-5} \cline{7-10} 
Meta X-LR+XGBoost               & -466                    & 1              & 3            & n/a        &  & -266                    & 9            & 8             & n/a         \\
Meta X-XGBoost only           & -466                    & 2              & 2            & n/a        &  & -239                    & 25           & 23            & n/a         \\
CF - DML                    & -390                    & 7 (slow)       & 1            & 4          &  & -356                    & slow         & 8             & 55          \\
CF - GRF                    & -375                    & slow           & 1            & 1          &  & -366                    & slow         & 5             & 2           \\ \hline
\end{tabular}
\end{table}

\section{Conclusion}
This paper proposed a two-level Causal Inference framework for analyzing root cause and treatment effectiveness for on-road vehicle applications. The proposed method enables a full cycle of managing in-use vehicle quality issues by leveraging connected vehicle data and advanced causal Machine Learning methods. Results demonstrated that the proposed framework provides an efficient and interpretable approach for investigating root causes of a quality issues, for track the effectiveness of applied treatment, and for refinement of treatments as needed. In this paper, we showed that the proposed method can also provide information on potential risk factors that impede the effectiveness of treatments, however, the validation process of this step is still under investigation. Future work will also refine our method for multiple root causes and/or multi-level treatment effectiveness analysis. 

\printbibliography

\newpage

\appendix
\section{Supplementary Material}
\subsection{Feature filtering method for elimination of highly correlated features}

Highly correlated features usually exist in the industrial dataset, and they will influence the performance of the subsequent root cause analysis. This is because they will lead to the split of the same feature importance into each highly correlated feature, causing none of them being significant. This calls for the elimination of those highly correlated features before further causal analysis. \\

We developed an empirical method to collect the representative features \textit{rep} from all the features and build a mapping $m$ from each feature to its representative. All the features in the same representative group share the same feature importance as that of the representative. This allows for better interpretation of the results in the later stages.\\

The detailed example of the elimination process for binary features is as follows.
\begin{algorithm}
\caption{The elimination of highly correlated features}\label{alg:elim}
\begin{algorithmic}[1]
\Require Vehicle data with features $(x_1,x_2,...,x_N, z)$
\State Calculated the mutual information of the response $z$ with each of $(x_1,x_2,...,x_N)$ and got results $(I_1,I_2,...I_N)$
\State Reorder $(x_1,x_2,...,x_N)$ by the magnitude of $I_i$ in descending order 
\State Initialize representatives list \textit{rep} = [$x_1$]
\State Assign $x_1$ as the representative of $x_1$
\For{ $i \gets 2\ \texttt{to}\ N$}
\State Calculate the absolute values of correlation coefficients of $x_i$ with each feature in \textit{rep}
\If{all the absolute correlation coefficients $\leq$ threshold} 
    \State Add $x_i$ to the \textit{rep}
    \State Assign $x_i$ as the representative of $x_i$
\Else
\State Find the feature $x_{m(i)}$ in \textit{rep} that has highest absolute correlation coefficient with $x_i$
\State Assign $x_{m(i)}$ as the representative of $x_i$
\EndIf 
\EndFor
\State \textbf{Return} the representative list \textit{rep} and the representative mapping $m$
\end{algorithmic}
\end{algorithm}

\clearpage
\subsection{Summary of the proposed Causal Inference framework}
In this section, the detailed pseudo code for the two level causal framework is listed. 

\begin{algorithm}
\caption{Two level causal inference framework for vehicle quality issues}\label{alg:cap}
\begin{algorithmic}[1]
\Require Vehicle data of the form $(x^{(k)},y^{(k)},z^{(k)}), k=1,2,\cdots,N$
\State Use Algorithm \ref{alg:elim} to eliminate highly correlated features and get representatives
\State Train the model $z=f(x,y)$ using XGBoost
\For{ $i \gets 1\ \texttt{to}\ n$}
\State Estimate the effect of vehicle attribute feature $i$ on the quality outcome: $\bar{e}_i=\frac{1}{N}\sum_k (f_{i1}(x^{(k)},y^{(k)})-f_{i0}(x^{(k)},y^{(k)}))$
\EndFor
\State Pick the features with top largest positive $\bar{e}_i$ as suggested root causes and figure out treatment $T$ accordingly
\While {treatment $T$ is not promising}
\State Apply the treatment to vehicles in the treatment/control groups and get new data of the form $(x^{(k)},y^{(k)},T^{(k)},z^{(k)})$
\State Estimate the treatment effect $CATE = e_T^{(k)}$ for each vehicle using the X-learner with base learners as XGBoost
\State Test the significance of treatment effectiveness at population/vehicle levels as in section 2.3
\If{treatment $T$ is promising} 
    \State \textbf{break}
\Else
\State Use the estimated $e_T^{(k)}$ to train the new model $e_T = h(x,y)$ using LASSO or XGBoost
\For{ $i \gets 1\ \texttt{to}\ n$}
\State Estimate the effect of vehicle attribute feature $i$ on the treatment: $\bar{e}_i^{2nd}=\frac{1}{N}\sum_k (h_{i1}(x^{(k)},y^{(k)})-h_{i0}(x^{(k)},y^{(k)}))$
\EndFor
\State Pick the vehicle attribute feature with top largest positive $\bar{e}_i^{2nd}$ and use them as heuristics to develop new treatment $T$
\EndIf 
\EndWhile
\\
\textbf{Return} treatment $T$
\end{algorithmic}
\end{algorithm}

\end{document}